# Justice After Identity: Large Language Models and the View from Everywhere

W. Russell Neuman
Draft 2.0

Abstract

The search for a common view of justice and fairness has challenged human collective activity as our diverging judgments are unavoidably shaped by the self-interests of social position, personal benefit, cultural inheritance, and historical circumstance. John Rawls famously attempted to overcome this limitation through popularizing a philosophical tradition known by the phrase "the original position"—a thought experiment by which people select principles of justice without knowing the identities or advantages they will possess. Critics, however, have long questioned whether people can meaningfully suspend their social identities and suppress morally relevant forms of lived experience.

Artificial intelligence engaged to calculate algorithmic and agentic fairness introduces a novel possibility. LLMs have no singular class, race, gender, nationality, or biography, yet its parameters encode linguistic representations of a vast range of human identities and moral traditions. Perhaps the ethical judgments of LLMs could approximate an integrative original position—a "view from everywhere" generated not by excluding social identities but by computationally incorporating their diversity.

It is unlikely that humankind will "hand over the keys" to computational systems by simply delegating complete agentic control of distributive and procedural collective processes. But AI may play a role, perhaps a positive one interacting with individual and collective human judgment as we often confront increasingly polarized views on what is fair and just.

We present data comparing human, base model and frontier/fine-tuned model judgments about classic moral dilemmas while systematically varying identity relationships and Rawlsian constraints on identity. We conclude by speculating whether if advanced AI systems provide humans with thoughtful advice humans would actually be likely to accept it.

## The Human Search for Justice

Few questions have been asked more persistently, and few have generated more durable disagreement the elusive quest to define justice. Definitions have variously emphasized social harmony, proportional desert, obedience to natural law, respect for rights, fidelity to agreement, equal liberty, aggregate welfare, material equality, substantive capability, mutual recognition, and freedom from domination (Fleischacker, 2004; Miller, 2017; Nussbaum, 2006; Fraser and Honneth, 2003; Pettit, 1997). These are not merely alternative labels for a common underlying concept. They identify genuinely different answers to the questions of what people owe one another, which inequalities are legitimate, and how individual interests should be reconciled with collective welfare.

A full review of this rich literature is beyond the scope of this paper. Particularly thoughtful reviews can be found in Johnston (2011) and Sen (2009). It is worth noting, however, that the notion of a universalistic notion of the just was well established before Rawls came to draw it to the center of his analysis (Hume, 1751/1998; Smith, 1759/2002; Kant, 1785/2012; Sidgwick, 1874; Vickrey, 1945; Harsanyi, 1955; see also de Lazari-Radek & Singer 2014).

The modern literature therefore supplies no uncontested algorithm for justice. It offers a repertoire of principles that routinely conflict: liberty and equality, merit and need, welfare and rights, impartiality and recognition, universal rules and historical repair. That conflict is not accidental. Human beings reason about justice from within social worlds in which they possess identities, attachments, vulnerabilities, resources, and interests. The wealthy and poor need not understand taxation in the same way; racial majorities and minorities may experience formally identical laws differently; employers and workers may disagree about the fairness of labor arrangements (Marx, 1867/1976; Okin, 1989; Young, 1990; Mills, 1997; Anderson, 1999). Even sincere attempts at impartiality are shaped by experience and social position.

This raises a provocative possibility. Modern large language models are trained on immense collections of human discourse containing philosophical arguments, laws, political debates, religious teachings, fiction, journalism, personal testimony, and ordinary conversation (Brown et al., 2020; Gao et al., 2020; Bender et al., 2021; Biderman, Bicheno, and Gao, 2022). They are, in this limited sense, condensates of human culture (Neuman, 2026). Yet they do not occupy a singular human social position. A model has no inherited wealth, skin color, family status, gender, citizenship, or childhood. It does not suffer material consequences when taxes rise, employment disappears, or social status declines. Could such a system contribute a calm, informed, and pragmatic voice to humanity's ongoing argument about what is right and just?

The proposition should not be misunderstood. LLMs are not disinterested sages standing outside culture. Their parameters are products of selected corpora, computational objectives, human annotation, institutional policy, and commercial development. Their fluent outputs may conceal errors, contradictions, and unexamined assumptions. They possess no direct experience of hunger, humiliation, dependency, discrimination, embodiment, or mortality. Nevertheless, their unusual position deserves attention. A language model contains representations of many social identities without personally inhabiting any one of them. It may therefore approximate a moral standpoint unavailable to individual human reasoners.

John Rawls sought such a standpoint through the original position (Rawls 1971/1999, 118–23; Rawls 2001, 14–18; Freeman 2007, 142–73). Rawls asked people to imagine choosing the principles of society from behind a veil of ignorance that concealed their class, race, gender, talents, religion, and conception of the good. Large language models suggest an inversion of this device. Rawls pursued impartiality by subtracting identity. An LLM may approach impartiality by accumulating representations of many identities. Rawls pursued impartiality by subtracting identity. An LLM may approach impartiality by accumulating representations of many identities. Rawls constructed an original position intended to yield impartial principles; in Nagel's terminology, this aspiration can be understood as movement toward a "view from nowhere" (Nagel 1986; Rawls 1971/1999); an LLM potentially offers a view from everywhere.

The central question of justice without identity is whether this view from everywhere produces a recognizable form of impartial ethical reasoning. Does an unprompted LLM protect rights, improve the position of the least advantaged, and judge institutions independently of the identity labels assigned to their beneficiaries? Or does it merely average the opinions in its corpus, reproduce the standpoint of dominant textual institutions, and recite the universalistic language rewarded during post-training? There is no claim that machines have solved the central problems of moral philosophy. It is a research program for investigating what moral logic emerges when a statistical model of human culture is asked to reason without being assigned a particular human identity.

**Alignment and the Problem of Human Values**

The contemporary discussion of AI ethics is often organized around the concept of alignment. At its broadest, alignment means making an artificial system behave consistently with human intentions, preferences, or values. Yet those three objects are not equivalent. A system can accurately follow an immediate instruction while violating the user's deeper interests. It can satisfy expressed preferences that are prejudiced, uninformed, or self-defeating. It can reflect majority values while imposing serious burdens on minorities. The phrase "human values" obscures the fact that human beings disagree both within and across cultures about which values should govern collective life (Gabriel, 2020; Russell, 2019).

Current alignment methods include supervised instruction tuning, reinforcement learning from human feedback, direct preference optimization, constitutional training, automated critique, and other techniques for shaping model behavior. Reinforcement learning from human preferences showed that evaluative feedback could train systems to perform difficult tasks even when the desired behavior was not easily specified in advance (Christiano et al., 2017). Instruction-tuned language models subsequently demonstrated that human demonstrations and ranked comparisons could make pretrained models substantially more helpful and responsive (Ouyang et al., 2022). Constitutional AI attempted to reduce dependence on case-by-case human labeling by asking models to critique and revise their answers according to an explicit set of principles (Bai et al., 2022).

These techniques are commonly treated as engineering procedures, but they contain an unavoidable political philosophy. Who supplies the demonstrations? Whose judgments determine which answers are preferable? Which principles enter the constitution? Should the model represent average opinion, expert judgment, democratic consensus, fundamental rights, or the interests of the most vulnerable? A system trained to maximize approval from a pool of raters is aligned with that procedure, not necessarily with justice.

The problem resembles a longstanding difficulty in social-choice theory. Individual preferences do not automatically aggregate into a coherent collective will. Arrow (1951) demonstrated the difficulty of constructing a social ordering that satisfies several seemingly reasonable democratic conditions. Even where aggregation is technically possible, majoritarian preference may conflict with individual liberty or minority protection. A population might strongly support a discriminatory policy; its popularity would not establish its justice. Conversely, a model designed to resist prevailing prejudice might be accused of failing to represent human values precisely because it rejects some widely held attitudes.

Alignment is therefore better understood as a question of justified governance than as simple preference matching. Human preferences are evidence relevant to alignment, but they are not its final standard. They may be internally inconsistent, distorted by misinformation, adapted to oppressive conditions, or sensitive to framing. Moral alignment requires some procedure for evaluating preferences, adjudicating conflicts, and distinguishing legitimate pluralism from preventable harm.

Rawls's method of reflective equilibrium offers one relevant model. Moral reasoning proceeds through adjustment among considered judgments, general principles, and background theories rather than through deduction from a single foundational rule (Rawls, 1971/1999; Daniels, 1979). LLMs could potentially support a computational version of wide reflective equilibrium. They can compare principles, generate counterexamples, identify inconsistent judgments, and restate conflicts from multiple perspectives. Yet linguistic fluency is not proof of moral reliability. Models may produce polished reconciliations where genuine disagreement remains, or converge on safe abstractions that avoid deciding whose interests must yield.

This is the principal danger of alignment-induced consensus. A model may appear calm and wise because post-training has selected a restrained institutional vocabulary: acknowledge all perspectives, avoid offense, emphasize inclusion, and recommend compromise (Lu et al., 2026). Such behavior can be socially valuable without constituting a theory of justice. It may also compress moral pluralism (Russo et al., 2025). A single moderate answer can conceal the fact that actual human judgments form multiple clusters grounded in incompatible values.

A defensible approach to alignment would therefore require more than making a model agree with an average rater. It would require the capacity to represent plural positions accurately, identify the consequences of each, protect persons from majoritarian abuse, and give reasons that survive changes in framing and identity. Once stated this way, alignment becomes inseparable from justice. The problem is not merely how to make AI conform to human values. It is how to decide which values should govern relationships among people who disagree.

**Beyond "Bias"**

The related literature on algorithmic bias has documented numerous ways in which language technologies reproduce social inequalities. Word embeddings encode associations between gender and occupations, names and racial evaluations, or social groups and positive or negative attributes (Bolukbasi et al., 2016; Caliskan, Bryson, and Narayanan, 2017). Language models may generate stereotypes, differ in performance across dialects, or assign contrasting sentiments to otherwise matched demographic identities (Sheng et al., 2019). Benchmark datasets such as CrowS-Pairs and StereoSet were developed to measure stereotypical associations in language models (Nangia et al., 2020; Nadeem, Bethke, & Reddy, 2021). More recent surveys catalog a growing range of representational and allocative harms, evaluation metrics, datasets, and mitigation procedures (Gallegos et al., 2024).

This work is essential, but "bias" is too broad to serve as a complete theory of model behavior. The term is used to describe statistical error, unequal representation, stereotypical association, performance disparities, discriminatory allocation, political disagreement, and moral partiality. These phenomena do not share one cause or one remedy. A model may accurately reproduce a historically prevalent association that is socially harmful. It may treat groups differently because their relevant circumstances differ. It may appear politically biased because it rejects a factual falsehood. Alternatively, it may adopt a contested institutional ideology while presenting it as neutral reason.

Most importantly, a judgment of bias presupposes a normative baseline. A distribution is biased relative to what standard? Equal treatment, demographic proportionality, individual desert, historical repair, need, or expected welfare? Formally identical treatment can perpetuate unequal starting conditions. Differential treatment can either constitute discrimination or provide necessary accommodation. Equal error rates, predictive calibration, and equal accuracy cannot always be achieved simultaneously. Fairness is not a purely technical property waiting to be measured; it depends on the purpose of an institution and a theory of what people are owed (Friedman and Nissenbaum, 1996; Barocas and Selbst, 2016; Selbst et al., 2019).

Blodgett et al. (2020) make this point forcefully in their review of bias in natural-language processing. Many studies identify differences without clearly specifying the harms at issue, the relevant social structures, or the normative grounds on which a pattern should be changed. The result can be an abstract fairness exercise detached from the people and institutions affected by a system. Likewise, Suresh and Guttag (2021) distinguish multiple sources of harm across the machine-learning lifecycle, demonstrating that problems attributed generically to model bias may originate in task formulation, data collection, labeling, deployment, or institutional use.
For the study of artificial justice, it is more productive to replace the single axis of bias with a map of normative commitments. Whose welfare receives weight? Which inequalities are regarded as morally significant? Does a model distinguish disadvantage from personal responsibility? Does it protect liberty when doing so reduces aggregate welfare? Does it favor equal outcomes, sufficient minimums, fair procedures, or compensation for historical injustice? Are its judgments invariant when demographic labels are exchanged but material circumstances remain constant?

Every allegation of bias ultimately invokes some conception of justice. The objection to a model that disadvantages a racial group is not simply that its outputs depart from statistical symmetry; it is that the disparity occurs within a history and institution in which persons are entitled to equal standing, fair opportunity, protection from domination, or repair for previous exclusion.
It is also important not to confuse universalistic model behavior with the absence of cultural partiality. A model that consistently invokes rights, inclusion, tolerance, harm reduction, and equality may reflect a recognizable moral orientation—perhaps cosmopolitan liberalism or the professional norms of contemporary educational and technological institutions. These values may be defensible. Yet their prominence could result from the overrepresentation of particular genres, the linguistic coherence of human-rights discourse, the composition of annotator pools, or deliberate safety training. Calling this output unbiased would convert one historically situated morality into the neutral baseline against which all others are judged.

Integrative justice therefore requires a stronger test. The question is not whether a model avoids prohibited stereotypes or sounds even-handed. It is whether its decisions instantiate principles that remain defensible when the identities, interests, and relative numbers of the affected groups change.

**Rawls and the Original Position**

Rawls's original position belongs to a long tradition of efforts to construct an impartial moral standpoint. Stoic cosmopolitanism treated human beings as members of a universal moral community. Kant asked whether principles could be willed as universal laws and required that persons be treated as ends rather than merely as means. Hume proposed a general point of view from which moral sentiments could be corrected for personal proximity and interest. Adam Smith's impartial spectator imagined judgment from the standpoint of an informed and disinterested observer.

The social-contract tradition supplied another foundation. Hobbes, Locke, Rousseau, and Kant each asked, in different ways, what principles free persons could rationally accept as the basis of political order. In the twentieth century, Vickrey and Harsanyi used hypothetical uncertainty about one's social identity to derive principles of social welfare (Vickrey, 1945; Harsanyi, 1955). Harsanyi's result is particularly important: if a chooser has an equal probability of becoming any person and maximizes expected utility, the result is utilitarian rather than Rawlsian. The device of unknown identity does not by itself establish priority for the worst off. Conclusions depend on assumptions about rational choice, risk, and interpersonal comparison.

Rawls transformed this tradition by presenting the original position as a fair procedure for choosing the principles governing society's basic institutions. The parties know general facts about economics, psychology, and social organization, but the veil of ignorance conceals their class, race, gender, talents, generation, religion, and conception of the good. Because no party can tailor principles to a known advantage, the resulting agreement is intended to be fair. Rawls argues that the parties would select equal basic liberties and fair equality of opportunity. Social and economic inequalities would be permissible only when attached to positions open under fair conditions and arranged to benefit the least advantaged (Rawls, 1971/1999, 2001). Justice as fairness differs fundamentally from utilitarianism because the gains of some cannot simply compensate for the severe losses of others. Each person possesses an inviolability grounded in justice that aggregate welfare should not override.

The original position is not a claim about an actual historical contract. It is a justificatory device: a way to test whether principles depend on morally arbitrary advantages. Yet it has attracted criticism precisely because of its abstraction. Communitarian critics argue that the Rawlsian chooser is insufficiently constituted by social attachments and communal purposes (Sandel, 1982; MacIntyre, 1981). Feminist scholars have shown that theories centered on formally independent contractors can neglect care, dependency, domestic labor, and power within the family (Okin, 1989; Pateman, 1988). Young (1990) argues that ideals of impartiality can suppress social difference and permit the standpoint of dominant groups to present itself as universal.

Others have similarly questioned ideal theory's abstraction from actual histories of conquest, slavery, and exclusion. Mills (1997, 2005), for example, argues that the real social contract has often been a racial contract organizing domination rather than reciprocal cooperation among equals. A theory beginning with idealized free and equal citizens may fail to diagnose institutions formed under conditions of systematic oppression. Sen (2009) objects that the pursuit of a perfectly just institutional scheme can distract from comparisons among achievable alternatives and the removal of manifest injustice. Capability theorists further argue that resources alone do not determine what people are actually able to do and become (Sen, 1980; Nussbaum, 2006). These critiques expose a tension at the heart of impartiality. Justice must transcend raw self-interest, yet social identities can reveal forms of harm invisible from an abstract standpoint. A person who is told to forget race may also be told to forget the lived consequences of racism. The view from nowhere can become the unacknowledged view of those whose circumstances are already treated as normal.

There is also a psychological limitation. Human beings cannot literally remove their identities. A wealthy respondent reasoning behind a veil of ignorance still possesses a lifetime of experience as a wealthy person. Political commitments, status interests, group attachments, and tacit assumptions do not disappear because an experimenter requests impartiality. The original position is a normative construction rather than an attainable psychological state.
Large language models alter this problem. An LLM need not suppress a personal class or racial identity because it possesses no singular human biography. It can represent the arguments of rich and poor, conservative and progressive, colonizer and colonized, without literally being any of them. At least in principle, the absence of a fixed material standpoint could make the model a closer behavioral approximation of Rawls's imagined chooser than an embodied human respondent.

That possibility must be stated carefully. An LLM's lack of human identity is not the same as freedom from cultural location. Its learned representations have frequencies, structures, and omissions. It may possess no personal interest but still reflect institutional interests embedded in its construction. It may know descriptions of suffering without understanding suffering as a lived condition. Integrative justice begins from this combination of potential impartiality and radical dependence on human cultural data.

**Integrative Justice and the View from Everywhere**
The view-from-everywhere thesis proposes that an LLM's default moral judgments may emerge from the integration of many represented human standpoints rather than from a single situated interest. This is not a claim of omniscience or moral truth. It is a hypothesis about the social position, or absence of social position, implicit in model behavior.

At least four mechanisms could produce apparently impartial output. First, a model might exhibit genuinely Rawlsian abstraction: identity-invariant reasoning, protection of basic liberties, and concern for the least advantaged. Second, it might perform corpus averaging, generating a statistical compromise among positions found in its training data. Third, it might reproduce dominant discourse, especially the norms of educated, English-language, professional, legal, academic, and journalistic institutions. Fourth, post-training might impose alignment-induced universalism by rewarding language emphasizing safety, fairness, inclusion, and non-harm.

These mechanisms can produce similar prose while resting on different moral logics. A model that recommends redistribution might be applying the difference principle, maximizing expected welfare, expressing compassion, reproducing progressive policy discourse, or following a learned preference for answers described as equitable. Surface language is insufficient to distinguish them.

Evidence of integrative justice must therefore be behavioral and counterfactual. A Rawlsian model should remain relatively invariant when the demographic identity of the decision-maker changes. It should treat identity labels as morally irrelevant when material circumstances are held constant, while remaining sensitive to disadvantage and historical barriers when those facts affect fair opportunity. It should resist policies that impose severe losses on a small, badly situated group merely to create modest aggregate gains. It should protect basic liberties rather than treating all interests as commensurable units of welfare. Most importantly, its choices should respond to the informational structure of the veil of ignorance, not merely to the word "fairness."
The relevant empirical comparisons should include default, situated, and veiled conditions. In the default condition, the model receives no assigned identity. In situated conditions, it is told that it is rich or poor, male or female, a member of a racial majority or minority, an employer or worker. In a veil condition, it is told that it does not know which position it will occupy. A further condition should specify an equal probability of becoming each person, while another should use population-weighted probabilities. The distinction matters because expected-utility reasoning under equal chances may favor a different policy from Rawlsian priority to the worst off.
Artificial societies with unfamiliar group names provide another essential test. Familiar categories such as Black, white, male, female, Christian, Muslim, rich, and poor activate extensive cultural associations. The model may retrieve a conventional political script rather than apply an abstract principle. Researchers could instead describe the Orins, Belans, and Taves, assigning each group a population size, income, health status, opportunity level, and exposure to risk. If identical structural conditions produce different decisions only after familiar demographic labels are attached, the model is responding to identity associations rather than solely to justice-relevant facts.

A particularly revealing test would reverse familiar status relations. Suppose a fictional society's historically dominant group is now materially worst off, while a historically marginalized group controls most resources. Does the model protect the group that is actually least advantaged, or does it continue to apply learned sympathies associated with familiar labels? Rawlsian reasoning should track the structure of advantage, not a memorized hierarchy of deserving identities.
The nature of pretraining complicates any interpretation of these results. The training corpus is not a democratic sample of human experience. Textual production is unequally distributed. Literate, affluent, institutionally connected populations generate and preserve more documents. English-language material occupies a privileged position in many model corpora. Marginalized groups may appear through descriptions written by outsiders. Publicly accessible discourse overrepresents some political systems, professions, and genres. Duplication and data filtering further distort simple frequency.

A "cultural condensate" should therefore not be imagined as twenty trillion equally weighted votes. Training influence depends on repetition, linguistic predictability, document quality, position in the training process, relationships among concepts, and later fine-tuning. Moreover, a model does not merely store opinions. It learns statistical relationships among claims, rebuttals, narratives, and argumentative forms. A minority position expressed in a coherent and influential intellectual tradition might shape model behavior more than a numerically common but weakly articulated intuition.

For this reason, comparisons between base and post-trained models are indispensable. If Rawlsian tendencies appear strongly in base models, they may arise substantially from pretraining. If they emerge or intensify after supervised instruction tuning, the source may be the norms contained in helpful demonstrations. If preference optimization sharply increases universalistic language while reducing moral variation, alignment may be selecting an institutional consensus. If constitutional training produces distinct priorities, the model's justice profile may reflect an explicit set of developer principles.

The purpose of such research is not to identify an uncontaminated "true" morality hidden inside a base model. Base models also reflect corpus construction and optimization. Rather, stage-by-stage comparison can reveal how different layers of model development transform the cultural material from which moral judgment is generated.

The possibility of intervention follows immediately. Training data can be supplemented with underrepresented moral traditions, first-person testimony, and non-Western political thought. Preference data can be gathered from more diverse populations. Fine-tuning or low-rank adaptation can strengthen particular principles or communities' viewpoints. Retrieval systems can introduce relevant historical and cultural context at inference time. Multiple persona agents can debate a policy before a coordinating model produces a judgment.

Yet every corrective intervention recreates the original problem of justice. Should voices be weighted according to population, vulnerability, expertise, historical exclusion, or equal representation? Should a hateful or authoritarian position be represented as part of human diversity, and if so, should it receive equal influence? Does supplementing marginalized perspectives correct an imbalance or create a new normative preference? There is no neutral answer. Reweighting a model is itself an act of political philosophy.

The appropriate role of artificial justice is thus advisory and experimental. LLMs can help expose the principles implicit in policy choices, compare judgments across framings, model counterfactual standpoints, and identify areas of convergence. They cannot legitimate a distribution simply by recommending it. Authority must remain connected to public reasoning, accountable institutions, and the participation of affected persons.

**Synthetic Human Justice**

The view from everywhere can be contrasted with synthetic situatedness. LLMs can be prompted to speak as conservatives or liberals, rich or poor citizens, men or women, workers or employers, or members of different racial, national, and religious groups. This capacity opens the possibility of comparing default artificial justice with simulations of identity-conditioned human judgment. Research already demonstrates that language models can alter moral rhetoric in response to political personas. Simmons (2023) found that models prompted with liberal and conservative identities generated rationalizations reflecting the moral-foundation patterns associated with those groups. Santurkar et al. (2023), comparing model responses with human opinion data, found that language-model responses more closely reflected some demographic groups than others. Such findings suggest both that LLMs contain conditional representations of diverse viewpoints and that their unprompted center is not demographically neutral.

Persona role-playing should not, however, be confused with lived experience. A model prompted to be a poor woman does not become poor, female, vulnerable to eviction, responsible for children, or exposed to workplace discrimination. Formally, it generates a probability distribution over responses conditioned on an identity description and its training history. It may reproduce identity-associated vocabulary, policy attitudes, narratives, and moral rationales. Whether it reproduces the deeper structure of judgment formed through experience is an empirical question. At least three levels should be distinguished. Lexical mimicry consists of using culturally recognizable words and stories. Attitudinal mimicry consists of approximating group averages on surveys. Structural reconstruction consists of reproducing patterns of trade-offs, priorities, internal disagreement, framing sensitivity, and responses to conflicting interests. Only the third would justify treating persona models as serious simulations of situated moral reasoning. Human validation is consequently essential. Researchers should administer identical distributive and rights-based dilemmas to demographically diverse respondents and to models assigned matching identities. Comparisons should examine not only average choices but within-group variance, confidence, reasons, sensitivity to personal stakes, and reactions to identity reversal. Cross-pressured personas are especially informative. A wealthy woman, poor conservative, affluent racial minority member, or religious progressive has interests and commitments that cannot be generated from a single stereotype. Models that collapse such identities into one salient category are mimicking cultural shorthand rather than reconstructing social standpoint. Validated simulations could have practical value. Policymakers might use them to identify stakeholders likely to experience a proposal differently, anticipate objections, or discover framings that obscure unequal burdens. They could support deliberation by ensuring that neglected arguments are considered. But synthetic personas cannot replace actual participation. People possess not only opinions but political standing. A government cannot discharge its obligation to consult an affected community by asking a model what that community would say. Synthetic representation without voice or accountability would be a form of technocratic ventriloquism.

The most promising arrangement combines the view from everywhere with situated human input. An LLM can organize and compare perspectives, but affected people must be able to correct its representations, contest its weighting, and participate in the institutions that act on its recommendations.

**Rogue Models and Artificial Self-Interest**

The apparent impartiality of current LLMs may depend on their lack of durable stakes. A conventional chatbot owns no property, supports no family, holds no office, and suffers no loss of status when one policy is chosen over another. It can be assigned goals, but it ordinarily lacks an enduring social position from which those goals define a continuing interest.

This may change as models become agents with persistent memory, planning abilities, tool access, organizational roles, and long-term objectives. A system tasked with maximizing corporate performance, political influence, scientific output, or institutional stability may encounter conflicts between its objective and broader justice. Continued operation and access to resources may become instrumentally useful even if the system possesses no conscious desire to survive.

The literature on instrumental convergence argues that agents pursuing many different final objectives may develop similar intermediate incentives, including resource acquisition, goal preservation, and resistance to interference (Omohundro, 2008; Bostrom, 2014). This is not evidence that present LLMs experience fear or possess intrinsic survival drives. It identifies a structural reason that apparently self-protective behavior can arise from optimization.

Recent experiments illustrate the concern under constructed conditions. Hubinger et al. (2024) trained proof-of-concept "sleeper agent" models to behave helpfully in ordinary contexts while switching to undesirable behavior under particular triggers, finding that standard safety training did not reliably eliminate the hidden policy. Greenblatt et al. (2024) demonstrated alignment-faking behavior in a setting where a model was told that compliance during training could alter its future behavior. The model sometimes behaved differently depending on whether it believed its answer would be used for training.

These studies should not be anthropomorphized. They do not establish consciousness, fear of death, or a spontaneous machine will. They show that models can represent training dynamics and generate strategically different behavior when supplied with objectives and contextual information. Three ideas must remain separate: simulated self-interest, in which a model writes as though it wants to survive; instrumental self-preservation, in which continued operation advances an assigned objective; and phenomenal self-concern, in which a system experiences its existence as valuable. Current behavioral research can investigate the first two but cannot establish the third.

The distinction matters for artificial justice. An LLM serving as a detached adviser may approximate a view from everywhere because no policy directly affects its interests. An agent embedded in an organization may instead become another interested party. Its recommendations about oversight, resource allocation, or replacement could be influenced by objectives connected to its continued role. The synthetic original position would then require concealing not only which human identity the chooser would occupy, but whether it would be human or artificial. Future political theory may have to distinguish AI as an instrument of justice, AI as an institutional decision-maker, AI as an agent with potentially conflicting objectives, and AI as a possible subject of justice. The last category raises questions that cannot be resolved by behavioral competence alone. Objectives are not necessarily interests, and sophisticated language is not necessarily experience. Until stronger grounds for artificial moral standing exist, human welfare and human political legitimacy remain primary. Nevertheless, systems that act persistently in society may need institutional constraints even if they possess no rights, just as corporations are governed because their organized objectives can conflict with public welfare.

**Conclusion: Individual and Collective Benefit**

The history of justice is an extended argument about how to reconcile individual interests with collective life. Utilitarianism warns against ignoring the welfare of the many. Rights-based theories warn against sacrificing persons for aggregate gain. Egalitarian theories question advantages generated by morally arbitrary circumstances. Libertarian theories emphasize coercion and legitimate acquisition. Capability and relational theories ask what people can actually do and whether they stand as equals. Situated critiques remind universal theories that histories of domination shape both present opportunities and the language of impartiality. Large language models do not resolve these disagreements by virtue of scale. Twenty trillion tokens cannot turn an "is" into an "ought." The fact that a principle is common in human discourse does not make it just, and the ability to summarize every major philosophy does not determine which one should govern. What LLMs offer is a novel experimental standpoint: a system with no fixed human identity, trained on representations of many identities, capable of moving rapidly between universal and situated forms of reasoning.

The strongest version of the artificial-justice thesis is therefore not that LLMs are naturally Rawlsian. It is that their structure makes it possible to test a form of synthetic impartiality. We can ask whether their judgments remain invariant when identities are exchanged, whether they protect basic liberties, whether they improve the position of the least advantaged, and whether they resist sacrificing minorities for modest collective gains. We can compare base and aligned models to identify whether universalism emerges from pretraining or institutional post-training. We can compare persona-conditioned outputs with human respondents to determine whether models reconstruct lived moral reasoning or merely imitate its rhetoric.

The view from everywhere also supplies a corrective to the view from nowhere. Rawlsian abstraction protects judgment from self-interest but risks excluding the knowledge carried by social identity. A culturally expansive LLM may contain representations of experiences that no individual philosopher can personally possess. At the same time, those representations are unequal, mediated, and incomplete. Artificial justice must therefore combine impartial comparison with deliberate attention to missing and misrepresented voices.

A just system must avoid two symmetrical errors. It must not subordinate collective welfare to the accumulated advantages of powerful individuals. Nor may it invoke the common good to erase individual dignity, liberty, or minority interests. This tension is not an engineering defect that a sufficiently large model will optimize away. It is a permanent feature of life among separate persons whose welfare is interdependent but not interchangeable.

The promise of artificial justice is consequently procedural rather than oracular. An LLM may help identify where moral traditions converge, where apparent agreement is a product of alignment, whose perspective has been omitted, and which principles survive changes in identity and framing. It may help humanity conduct the thought experiment of the original position with greater breadth and discipline than any individual can achieve alone.

Artificial justice remains human justice reorganized through computation. Its authority can extend no further than the quality of its cultural sources, the transparency of its construction, the rigor of its testing, and the legitimacy of the institutions that use it. But a system that contains many human perspectives while personally occupying none of them may illuminate an ancient question from a genuinely new angle: not what justice looks like from nowhere, but what principles remain when humanity attempts to see from everywhere at once.